\documentclass[runningheads]{llncs}
\usepackage[T1]{fontenc}
\usepackage{graphicx}
\usepackage{amsmath,amssymb,amsfonts}
\usepackage{bm}
\usepackage{algorithm}
\usepackage{algorithmicx}
\usepackage{algpseudocode}
\usepackage{booktabs}
\usepackage[colorlinks=true,linkcolor=blue,citecolor=blue,urlcolor=blue]{hyperref}
\usepackage[misc]{ifsym}

\newcommand{\figref}[1]{Fig.~\ref{#1}}
\newcommand{\tabref}[1]{Table~\ref{#1}}
\newcommand{\ie}{\textit{i.\,e.}~}

\newcommand{\etal}{\textit{et al.}~}

\begin{document}
\title{MoViT: Memorizing Vision Transformers for Medical Image Analysis}
\titlerunning{Memorizing Vision Transformer}
\author{
Yiqing Shen, 
Pengfei Guo, 
Jingpu Wu, 
Qianqi Huang, \\
Nhat Le,
Jinyuan Zhou,  
Shanshan Jiang,  
Mathias Unberath
}
\authorrunning{Y. Shen et al.}
\institute{
Johns Hopkins University\\
\email{
\{yshen92,unberath\}@jhu.edu
}, \email{sjiang21@jhmi.edu}
}

\maketitle              % typeset the header of the contribution

\begin{abstract}
The synergy of long-range dependencies from transformers and local representations of image content from convolutional neural networks (CNNs) has led to advanced architectures and increased performance for various medical image analysis tasks due to their complementary benefits.
However, compared with CNNs, transformers require considerably more training data, due to a larger number of parameters and an absence of inductive bias. 
The need for increasingly large datasets continues to be problematic, particularly in the context of medical imaging, where both annotation efforts and data protection result in limited data availability.
In this work, inspired by the human decision-making process of correlating new ``evidence'' with previously memorized ``experience'', we propose a Memorizing Vision Transformer (MoViT) to alleviate the need for large-scale datasets to successfully train and deploy transformer-based architectures.
MoViT leverages an external memory structure to cache history attention snapshots during the training stage.
To prevent overfitting, we incorporate an innovative memory update scheme, attention temporal moving average, to update the stored external memories with the historical moving average.
For inference speedup, we design a prototypical attention learning method to distill the external memory into smaller representative subsets.
We evaluate our method on a public histology image dataset and an in-house MRI dataset, demonstrating that MoViT applied to varied medical image analysis tasks, can outperform vanilla transformer models across varied data regimes, especially in cases where only a small amount of annotated data is available. 
More importantly, MoViT can reach a competitive performance of ViT with only 3.0\% of the training data.
In conclusion, MoViT provides a simple plug-in for transformer architectures which may contribute to reducing the training data needed to achieve acceptable models for a broad range of medical image analysis tasks.

\end{abstract}

\section{Introduction}

With the advent of Vision Transformer (ViT), transformers have gained increasing popularity in the field of medical image analysis~\cite{vit}, due to the capability of capturing long-range dependencies.
However, ViT and its variants require considerably larger dataset sizes to achieve competitive results with convolutional neural networks (CNNs), due to larger model sizes and the absence of convolutional inductive bias~\cite{vit,liu2021efficient}.
Indeed, ViT performs worse than ResNet~\cite{resnet}, a model of similar capacity, on the ImageNet benchmark\cite{imagenet}, if ViT does not enjoy pre-training on JFT-300M~\cite{jft300m}, a large-scale dataset with 303 million weakly annotated natural images.
The drawback of requiring exceptionally large datasets prevents transformer-based architectures to fully evolve its potential in the medical image analysis context, where data collection and annotation continue to pose considerable challenges.
To capitalize on the benefits of transformer-based architectures for medical image analysis, we seek to develop an effective ViT framework capable of performing competitively even when only comparably small data is available.

In the literature, the problematic requirement for large data is partly alleviated by extra supervisory signals. Data-efficient Image Transformer (DeiT), for example, distills hard labels from a strong teacher transformer~\cite{touvron2021training}. 
Unfortunately, this approach only applies to problems where data-costly, high-capacity teacher transformer can be developed. 
Moreover, DeiT enables the training on student transformers exclusively for mid-size datasets, between 10k to 100k samples, and the performance dramatically declines when the data scale is small~\cite{touvron2021training}.
Concurrently, another line of work attempts to introduce the shift, scale, and distortion invariance properties from CNNs to transformers, resulting in a series of hybrid architecture designs~\cite{li2021localvit,wu2021cvt,xu2021co,yuan2021incorporating}.
To give a few examples, Van \etal fed the extracted features from CNNs into a transformer for multi-view fusion in COVID diagnosis~\cite{tulder2021multi}.
Barhoumi \etal extended a single CNN to multiple CNNs for feature extraction before the fusion by a tansformer~\cite{barhoumi2021scopeformer}.
Importantly, they note that pre-training on ImageNet is still required to fuse convolutional operations with self-attention mechanisms, particularly in the medical context~\cite{barhoumi2021scopeformer}.
Yet, pre-training on large-scale medical dataset is practically unaffordable, due to the absence of centralized dataset as well as the privacy regularizations. 

Our developments to combat the need for large data to train transformer-models is loosely inspired by the process that clinicians use when learning how to diagnose medical images from a relative very limited number of cases compared to regular data size.
To mimic this human decision-making process, where new information or ``evidence'' is often conceptually correlated with previously memorized facts or ``experience'', we present the \textit{Memorizing Vision Transformer} (MoViT) for efficient medical image analysis.
MoViT introduces external memory, allowing the transformer to access previously memorized experience, \ie keys and values, in the self-attention heads generated during the training.
In the inference stage, the external memory then enhances the instance-level attention by looking up the correlated memorized facts.
Introducing external memory enables long-range context to be captured through attention similar to language modeling, which provides supplementary attention with the current ViT and variants~\cite{mem_vit,khandelwal2019generalization,guo2022learning}.

The contributions of this paper are three-fold, summarized as follows. 
(1) A novel \textit{Memorizing Vision Transformer} (MoViT), which introduces storage for past attention cues by caching them into external memory, without introducing additional trainable parameters.
(2) A new approach to updating the memory using a \textit{Attention Temporal Moving Average} scheme, that accumulates attention snapshots and optimizes data in the external memory dynamically. In contrast, previous work, such as \cite{mem_vit}, is restricted to a random dropping-out scheme to keep a fixed amount of external memorized events.
(3) A new \textit{post hoc} scheme, \textit{Prototypical Attention Learning}, to distill the large-scale cached data into a representative prototypical subset, which accelerates computation during inference.
Experiments are carried out across different modalities, \ie Magnetic Resonance (MRI) Images and histopathological images, demonstrating superior performance to vanilla transformer models across all data regimes, especially when only small amounts of training samples are available.

\section{Methods}

\begin{figure}[t!]
    \centering
    \includegraphics[width=0.95\linewidth]{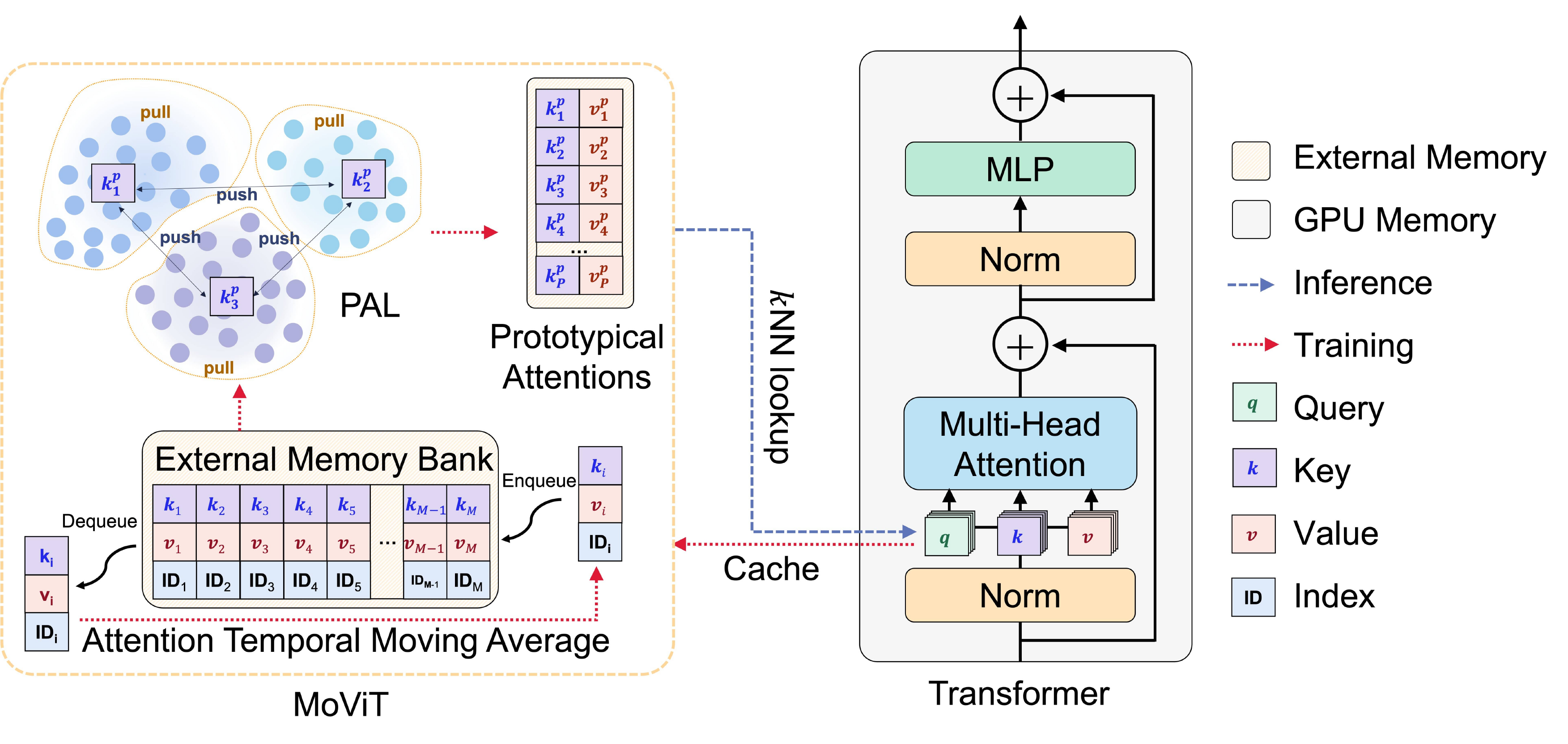}
    \caption{An overview of the proposed Memorizing Vision Transformer (MoViT). 
    }
    \label{fig:overview}
\end{figure}

\subsubsection{Memorizing Vision Transformer.} 
Memorizing Vision Transformer (MoViT) accumulates snapshots of attention cues \ie key $k$ and value $v$ generated by the attention heads as the memorized ``experience'', and caches them to an external memory bank in the form of an indexed triplet ($ID$, $k$, $v$) during the training process.
The enumerated index $ID$ of the data sample and the generated attention fact ($k,v$) are prepared for an efficient lookup and update in the subsequent Attention Temporal Moving Average (ATMA) scheme.
Importantly, the gradients are not back-propagated into the memory bank, and thus, the caching operation only costs slightly extra training.
This approach is practical in that the proposed MoViT can be easily plugged into any Vision Transformer (ViT) or its variants, by replacing one or multiple vanilla transformer blocks with MoViT blocks.
An overview of the MoViT framework is presented in \figref{fig:overview}.

\subsubsection{Attention Temporal Moving Average.} 
To remove stale memorized ``experience''~\cite{wang2020cross}, and also to prevent overfitting to the training set, we introduce a novel Attention Temporal Moving Average (ATMA) strategy to update external memory.
Current approaches routinely employ a fixed capacity to store triplets, where outdated cached memory triplets are dropped and new facts are taken in randomly.
Different from this approach, we improve the mechanism by accumulating all the past snapshots \textit{w.r.t} index $ID$, with the introduction of Exponential Moving Average update~\cite{meanteacher}.
The outdated cached ``experience'' $(k_\text{old},v_\text{old})$ \textit{w.r.t} index $ID$ denote the fact generated in the previous epoch, and is updated by the subsequently generated facts $(k_\text{generated},v_\text{generated})$ in the proposed ATMA according to:
\begin{equation}
\begin{cases}
\begin{aligned}
    k_{\text{new}} & = \alpha_k \cdot k_{\text{generated}} + (1-\alpha_k) \cdot k_{\text{old}}, \\
    v_{\text{new}} & = \alpha_v \cdot v_{\text{generated}} + (1-\alpha_v) \cdot v_{\text{old}},
\end{aligned}
\end{cases}
\end{equation}
where the subscripts ``new'' denotes the updated facts to the external memory, and $\alpha_k$, $\alpha_v$ are the friction terms.
In the smoothing process, both coefficients uniformly follow the ramp-down scheme~\cite{ramp} for a steady update. 
Specifically, the coefficients are subject to the current training epoch number $t$, \ie
\begin{equation}
  \alpha = 
  \begin{cases}
 1 - \alpha_0 \cdot \exp (-t_0(1 - \frac{t}{t_0})^2, &t \le t_0 \\
 1 - \alpha_0, &t > t_0 \\
  \end{cases}
\end{equation}
with $a_0 = 0.01$ and $t_0$ set to 10\% of the total training epochs as in previous work~\cite{meanteacher}.
The number of stored facts $M$ is exclusively correlated with the dataset scale, and network architecture \ie $M=\#(\text{training samples})\times\#(\text{attention heads})$, where the number of attention heads is often empirically set between the range of 3-12~\cite{vit}, leading to a bounded $M$.

\subsubsection{Prototypical Attention Learning.}  
We write all the cached experience from the training stage of MoViT as $\mathcal{F}=\{(k_i,v_i)\}_{i=1}^M$. 
Then, prototypical attention facts refer to a small number of representative facts to describe $\mathcal{F}$ \ie $\mathcal{P}=\{(k_i^p,v_i^p)\}_{i=1}^P$, where $P$ represents the total number of prototypes. 
To distill the external memorized facts into representative prototypes for efficient inference, we introduce Prototypical Attention Learning (PAL), which is applied \textit{post hoc} to the external memory after model training. 
To identify the prototype keys from the cached keys $\{k_i\}_{i=1}^M$, we leverage the Maximum Mean Discrepancy (MMD) metric~\cite{mmd} to measure the discrepancy between two distributions.
Subsequently, the objective in PAL is steered toward minimizing the MMD metric, \ie
\begin{equation}
% \argmin_{k_i^p} \Big(
MMD^2 = \frac{1}{P^2}\sum_{i,j=1}^P D(k_i^p,k_j^p) - 
\frac{1}{PM}\sum_{i,j=1}^{P,M}D(k_i^p,k_j) + 
\frac{1}{M^2}\sum_{i,j=1}^MD(k_i,k_j)
% \Big)
,
\label{eq:mmd}
\end{equation}
where $D(\cdot,\cdot)$ denotes the cosine similarity. 
We employ a greedy search to find $\{k_i^p\}_{i=1}^P$ from Eq. \eqref{eq:mmd}. 
To integrate all information after deriving prototype keys, we leverage the weighted average to derive the associated $\{v_i^p\}_{i=1}^P$ \ie
\begin{equation}
    v_i^p = \sum_{j=1}^M w_{j,i} v_j~~\text{with}~ w_{j,i}=\frac{\exp(D(v_j,v_i^p)/\tau)}{\sum \exp_{k=1}^M (D(v_k,v_i^p)/\tau)},\label{eq:avg}
\end{equation}
where the weights $w_{j,i}$ are normalized by the \texttt{softmax} operation, and temperature $\tau$ is a hyper-parameter to determine the confidence of normalization.

\subsubsection{Inference Stage.} 
To apply the attention facts $\mathcal{F}$ or prototypes $\mathcal{P}$ stored in the external memory during inference, approximate $k$-nearest-neighbor (kNN) search is employed to look up the top $k$ pairs of (key, value) \textit{w.r.t} the given the local queries. 
In this way, the same batch of queries generated from the test sample is used for both the multi-head self attentions and external memory retrievals.
With the retrieved keys, the attention matrix is derived by computing the \texttt{softmax} operated dot product with each query.
Afterwards, we use the attention matrix to compute a weighted sum over the retrieved values.
The results attended to local context and external memories are combined using a learned gate scheme~\cite{mem_vit}.

\section{Experiments}

\subsubsection{Datasets.} 
Evaluations are performed on two datasets curated from different modalities.
(1) 
Histology Image Dataset: \textit{NCT-CRC-HE-100K} is a public Hematoxylin \& Eosin (H\&E) stained histology image dataset with $100,000$ patches without overlap, curated from $N=86$ colorectal cancer samples~\cite{dataset}.
All RGB images are scaled to $224\times224$ pixels at the magnification of $20\times$. 
To simulate various data availability conditions, some experiments use a subset from \textit{NCT-CRC-HE-100K} as the training set.
In terms of the test set, an external public dataset \textit{CRC-VAL-HE-7K} with 7180 patches from $N=50$ patients is employed.
This dataset was designed to classify the nine tissue categories from histology image patches, and we use the top-1 test accuracy as the evaluation metric.
(2) 
MRI Dataset: This in-house dataset includes $147$ scans with malignant gliomas curated from $N=92$ patients.
All data has been deidentified properly to comply with the Institutional Review Board (IRB). 
Each scan contains five MRI sequences, namely T1-weighted (T1w), T2-weighted (T2w), fluid-attenuated inversion recovery (FLAIR), gadolinium enhanced T1-weighted (Gd-T1w), and amide proton transfer-weighted (APTw). 
The corresponding slices from each scan are concentrated after co-registration and z-score normalization, resulting in an input size of $256\times256\times5$.
A proportion of 80\% of the patients are divided into the training set, and the remaining 20\% as the test set \ie, 1770 training samples and 435 test samples.
This is a binary classification task to distinguish malignant gliomas from normal tissue.
We use accuracy, area under the ROC curve (AUC), precision, recall, and F1-score as the evaluation metrics.

\begin{figure}[t!]
    \centering
    \includegraphics[width=0.65\linewidth]{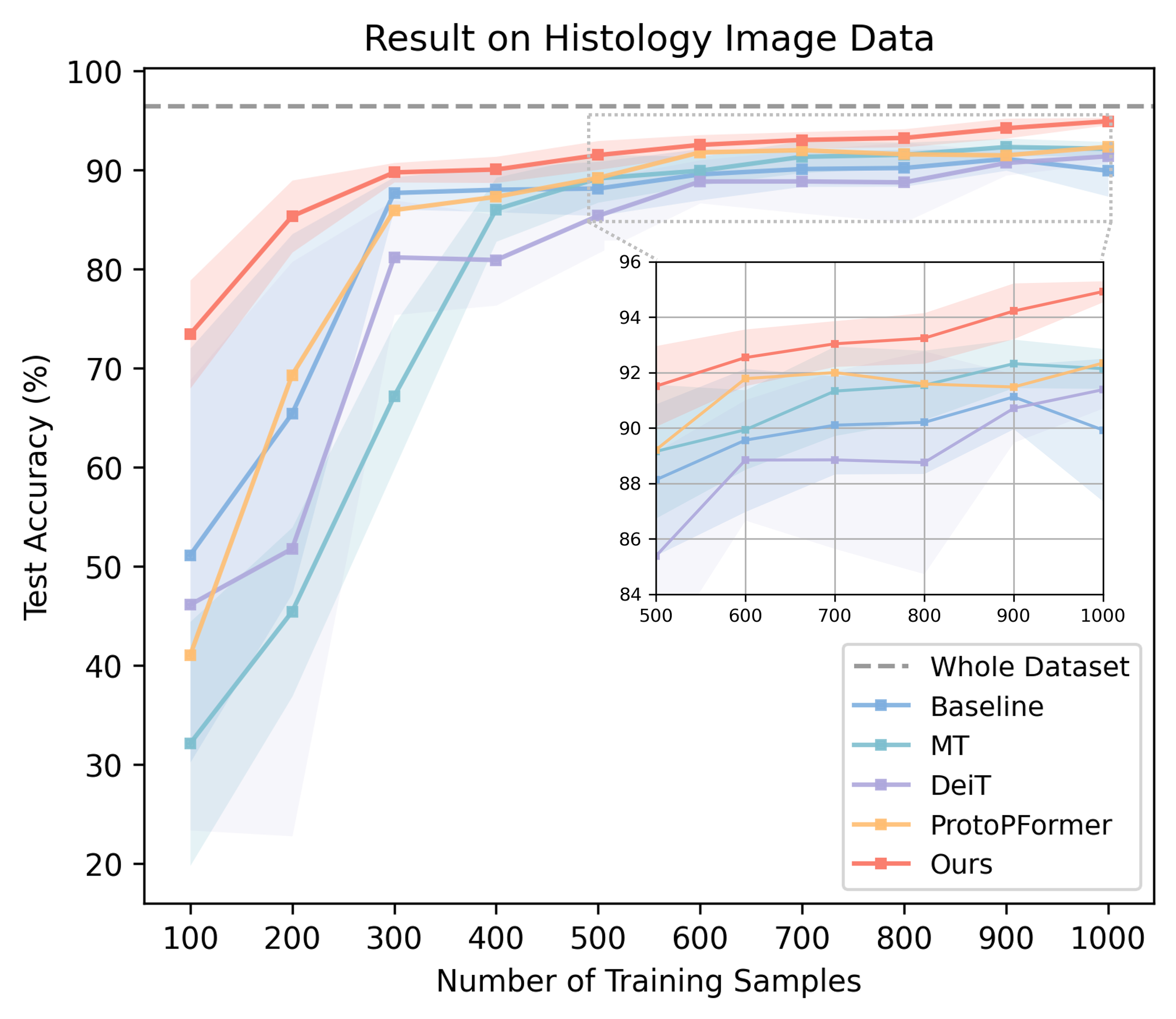}
    \caption{Performance comparisons of MoViT plugged into the last layer of ViT-Tiny with counterparts, using a varying number of training samples from the Histology image dataset ($0.1\%$ to $1\%$ data employed).
    The solid line and shadow regions represent the average and the standard deviation of test accuracy computed from seven random runs, respectively.
    The dashed line denotes the performance of the baseline (vanilla ViT) trained with the entire dataset, regarded as the performance upper bound.
    }
    \label{fig:result1}
\end{figure}

\begin{table}[hb!]
\caption{Performance comparison trained on the entire dataset in terms of test accuracy (\%). 
We employ three ViT configurations \ie ViT-Tiny, ViT-Small, and ViT-Base.} \label{tab:result1}
\begin{center}   
% \resizebox{\linewidth}{!}{ 
\begin{tabular}{l|ccc} 
% \hline
\toprule
Method & ViT-Tiny & ViT-Small & ViT-Base \\
\midrule
Baseline & $96.462$\tiny{$\pm0.213$} & $95.850$\tiny{$\pm0.503$} &  $94.231$\tiny{$\pm0.511$}\\
MT \cite{mem_vit} & $96.511$\tiny{$\pm0.312$} & $96.621$\tiny{$\pm0.108$} & $95.102$\tiny{$\pm0.272$}\\
DeiT \cite{touvron2021training} & $96.439$\tiny{$\pm0.331$} & $96.216$\tiny{$\pm0.213$} & $93.246$\tiny{$\pm0.259$} \\
ProtoPFormer \cite{xue2022protopformer} & $96.712$\tiny{$\pm0.521$} & $96.032$\tiny{$\pm0.364$} & $93.002$\tiny{$\pm0.752$}\\
\hline
MoViT (Ours) & \bm{$97.792$}\tiny{$\pm0.293$} & \bm{$97.326$}\tiny{$\pm0.138$}& \bm{$95.989$}\tiny{$\pm0.205$}\\
\bottomrule
\end{tabular}
% }
\end{center}
\end{table}

\subsubsection{Implementations.} 
All experiments are performed on one NVIDIA GeForce RTX 3090 GPU with 24 GB memory. 
An AdamW optimizer is used with a Cosine Annealing learning rate scheduler, where the initial learning rates are $2\times10^{-3}$ for the MRI dataset, $5\times10^{-4}$ for the Histology image dataset; with the maximum number of training epochs set to $100$.
We plug the MoViT into the last layer of ViT-Tiny, \ie 12 transformer layers with 3, and set $k=32$ for the kNN lookup.
In PAL, we set the number of prototypes $P=\#(\text{number of class})\times 32$, and temperature $\tau=0.5$ for Eq.~\eqref{eq:avg}.
Comparisons are made to Memorizing Transformer (MT)~\cite{mem_vit}, DeiT~\cite{touvron2021training}, and ProtoPFormer \ie a prototypical part network framework for ViT~\cite{xue2022protopformer}, where the vanilla ViT is regarded as the baseline.
To compute the mean and standard deviation of the metrics, all models are trained from scratch for seven random runs.

\begin{table}[b!]
\caption{Quantitative comparison on the MRI dataset.
MoViT achieves the highest performance across all metrics, further suggesting its ability to perform well in applications where limited data is available.
} \label{tab:result2}
\begin{center}   
% \resizebox{\linewidth}{!}{ 
\begin{tabular}{l|cccccc} 
% \hline
\toprule
Method & Accuracy($\uparrow$) & AUC($\uparrow$) & Precision($\uparrow$) & Recall($\uparrow$) & F1-score($\uparrow$) \\
\midrule
Baseline & $74.01$\tiny{$\pm0.40$} & $80.62$\tiny{$\pm0.40$} & $57.64$\tiny{$\pm0.66$} & $79.01$\tiny{$\pm0.68$} & $66.64$\tiny{$\pm0.54$} \\
MT \cite{mem_vit} & $77.92$\tiny{$\pm0.36$} & $84.83$\tiny{$\pm0.40$} & $62.54$\tiny{$\pm0.68$} & $81.84$\tiny{$\pm0.64$} & $70.95$\tiny{$\pm0.56$} \\
DeiT \cite{touvron2021training}& $78.63$\tiny{$\pm0.40$} & $85.65$\tiny{$\pm0.39$} & $63.07$\tiny{$\pm0.67$} & $84.68$\tiny{$\pm0.56$} & $72.20$\tiny{$\pm0.54$}\\
ProtoPFormer \cite{xue2022protopformer} &
$77.53$\tiny{$\pm0.40$} & $85.74$\tiny{$\pm0.38$} & $61.12$\tiny{$\pm0.67$} & $86.07$\tiny{$\pm0.53$} & $71.54$\tiny{$\pm0.55$} \\
\hline
Ours & \bm{$82.05$}\tiny{$\pm0.36$} & \bm{$88.38$}\tiny{$\pm0.30$} & \bm{$65.94$}\tiny{$\pm0.66$} & \bm{$94.43$}\tiny{$\pm0.36$} & \bm{$77.67$}\tiny{$\pm0.47$}\\
\bottomrule
\end{tabular}
% }
\end{center}
\end{table}

\subsubsection{Results on Histology Image Dataset.} 
To simulate the case where only small data is available as in many medical image analysis tasks, we use a limited proportion of the training set, across varied data regimes, and use the whole \textit{NCT-CRC-HE-100K} as the test set for a fair comparison.
As shown in \figref{fig:result1}, MoViT improves over the baseline at any data scale, especially when the number of samples is particularly small, \ie $0.1\%$, where we can observe a similar trend with a large proportion of the data between 1\%-100\%.
Notably, our method can achieve a close margin to the entire-dataset-trained model ($96.462\%\pm$0.213\%) using only 1.0\% data ($94.927\%\pm$0.378\%), and a competitive performance ($96.341\%\pm$0.201\%) with 3.0\% data.
Additionally, our approach also significantly reduces the performance fluctuations \ie standard deviation, leading to a more stable performance.
For example, vanilla ViT is $20.901\%$ when trained with $0.1\%$ data and ours is $5.452\%$ \ie approximately four times smaller.
Moreover, our method can consistently outperform state-of-the-art data-efficient transformer DeiT~\cite{touvron2021training} and pure prototype learning method ProtoPFormer~\cite{xue2022protopformer}.
We notice that Memorizing Transformer (MT) \cite{mem_vit} performs worse than the baseline although achieving almost 100\% training accuracy, where the gap becomes significant with $0.1\%-0.4\%$ data, which we attribute to the overfitting issue. 
The large margin between the performance of MT and MoViT implies that ATMA and PAL can alleviate the overfitting issues during the memorization of the facts.  
Performance comparison is also performed on the entire training set \ie using $100\%$ \textit{NCT-CRC-HE-100K} as the training set, with different ViT configurations. 
In \tabref{tab:result1}, our method can consistently outperform its counterparts with a large margin, which demonstrates its applicability and scalability to large datasets.
This suggests that MoViT scales well to a wide range of data scales as a by-product.
The averaged training times per epoch on ViT-Tiny are 162.61(s) for baseline ViT, 172.22(s) for MT, 109.8(s) for DeiT, 639.4(s) for ProtoPFormer, and 171.49(s) for our approach.
Our method can boost performance with a reduced training data scale.

\subsubsection{Results on MRI Dataset.} 
As depicted in~\tabref{tab:result2}, our proposed MoViT achieves the highest performance in terms of all metrics on the MRI dataset, where the dataset scale is relatively small, by nature.
Specifically, MoViT can improve the AUC by a margin of $0.026$ to the state-of-the-art transformer \ie $0.857$ achieved by ProtoPFromer; and can achieve better performance (AUC of $0.821$) than baseline with 30\% training data.
Empirically, our method is superior to other modalities in the generalization ability.

\subsubsection{Ablation Study.}
To investigate the contribution of each functional block, ablation studies are performed on the MRI dataset. 
As shown in~\tabref{tab:result3}, the proposed MoViT benefits from both ATMA and PAL. 
Although each module brings a similar AUC improvement from $0.010$ to $0.013$, the exclusion of the two modules suffers an AUC decline of $0.022$.
Conclusively, the reported results suggest the effectiveness and indispensability of ATMA and PAL.

\begin{table}[t!]
\caption{Ablations on MRI dataset with MoViT-Tiny as the backbone. We can observe that the exclusion of either ATMA or PAL results in decreased performance with varying degrees.
} \label{tab:result3}
\begin{center}   
% \resizebox{\linewidth}{!}{ 
\begin{tabular}{ll|cccccc} 
% \hline
\toprule
ATMA & PAL & Accuracy($\uparrow$) & AUC($\uparrow$) & Precision($\uparrow$) & Recall($\uparrow$) & F1-score($\uparrow$) \\
\midrule
 &  & $78.63$\tiny{$\pm0.39$} & $86.14$\tiny{$\pm0.38$} & $62.75$\tiny{$\pm0.68$} & $86.07$\tiny{$\pm0.60$} & $72.62$\tiny{$\pm0.54$}\\
 &  \checkmark  & $79.35$\tiny{$\pm0.34$} & $87.43$\tiny{$\pm0.31$} & $61.92$\tiny{$\pm0.63$} & $96.55$\tiny{$\pm0.31$} & $75.57$\tiny{$\pm0.49$}\\
\checkmark  &  &  $79.54$\tiny{$\pm0.38$} & $87.13$\tiny{$\pm0.32$} & $62.27$\tiny{$\pm0.66$} & $95.85$\tiny{$\pm0.31$} & $75.49$\tiny{$\pm0.48$} \\
\checkmark & \checkmark & $82.05$\tiny{$\pm0.36$} & $88.38$\tiny{$\pm0.30$} & $65.94$\tiny{$\pm0.66$} & $94.43$\tiny{$\pm0.36$} & $77.67$\tiny{$\pm0.47$} \\
\bottomrule
\end{tabular}
% }
\end{center}
\end{table}

\section{Conclusion}
In conclusion, we show that using memory in transformer architectures is beneficial for reducing the amount of training data needed to train generalizable transformer models. The reduction in data needs is particularly appealing in medical image analysis, where large-scale data continues to pose challenges. 
Our model, the Memorizing Vision Transformer (MoViT) for medical image analysis, caches and updates relevant key and value pairs during training. It then uses them to enrich the attention context for the inference stage. 
MoViT's implementation is straightforward and can easily be plugged into various transformer models to achieve performance competitive to vanilla ViT with much less training data.
Consequently, our method has the potential to benefit a broad range of applications in the medical image analysis context. 
Future work includes a hybrid of MoViT with convolutional neural networks for more comprehensive feature extraction. 

\subsubsection{Acknowledgments:}
This work was supported in part by grants from the National Institutes of Health (R37CA248077, R01CA228188). 
The MRI equipment in this study was funded by the NIH grant: 1S10ODO21648.

\bibliographystyle{splncs04}
\bibliography{6-ref}

\begin{thebibliography}{10}
\providecommand{\url}[1]{\texttt{#1}}
\providecommand{\urlprefix}{URL }
\providecommand{\doi}[1]{https://doi.org/#1}

\bibitem{barhoumi2021scopeformer}
Barhoumi, Y., et~al.: Scopeformer: n-cnn-vit hybrid model for intracranial hemorrhage classification. arXiv preprint arXiv:2107.04575  (2021)

\bibitem{vit}
Dosovitskiy, A., et~al.: An image is worth 16x16 words: Transformers for image recognition at scale. arXiv preprint arXiv:2010.11929  (2020)

\bibitem{guo2022learning}
Guo, P., et~al.: Learning-based analysis of amide proton transfer-weighted mri to identify true progression in glioma patients. NeuroImage: Clinical  \textbf{35},  103121 (2022)

\bibitem{resnet}
He, K., et~al.: Deep residual learning for image recognition. In: Proceedings of the IEEE conference on CVPR. pp. 770--778 (2016)

\bibitem{dataset}
Kather, J.N., et~al.: {100,000 histological images of human colorectal cancer and healthy tissue}. Zenodo  (Apr 2018). \doi{10.5281/zenodo.1214456}

\bibitem{khandelwal2019generalization}
Khandelwal, U., et~al.: Generalization through memorization: Nearest neighbor language models. arXiv preprint arXiv:1911.00172  (2019)

\bibitem{mmd}
Kim, B., et~al.: Examples are not enough, learn to criticize! criticism for interpretability. Advances in neural information processing systems  \textbf{29} (2016)

\bibitem{ramp}
Laine, S., et~al.: Temporal ensembling for semi-supervised learning. arXiv preprint arXiv:1610.02242  (2016)

\bibitem{li2021localvit}
Li, Y., et~al.: Localvit: Bringing locality to vision transformers. arXiv preprint arXiv:2104.05707  (2021)

\bibitem{liu2021efficient}
Liu, Y., et~al.: Efficient training of visual transformers with small datasets. Advances in Neural Information Processing Systems  \textbf{34},  23818--23830 (2021)

\bibitem{imagenet}
Russakovsky, O., et~al.: Imagenet large scale visual recognition challenge. International journal of computer vision  \textbf{115}(3),  211--252 (2015)

\bibitem{jft300m}
Sun, C., et~al.: Revisiting unreasonable effectiveness of data in deep learning era. In: Proceedings of the ICCV. pp. 843--852 (2017)

\bibitem{meanteacher}
Tarvainen, A., et~al.: Mean teachers are better role models: Weight-averaged consistency targets improve semi-supervised deep learning results. Advances in neural information processing systems  \textbf{30} (2017)

\bibitem{touvron2021training}
Touvron, H., et~al.: Training data-efficient image transformers \& distillation through attention. In: International Conference on Machine Learning. pp. 10347--10357. PMLR (2021)

\bibitem{tulder2021multi}
Tulder, G.v., et~al.: Multi-view analysis of unregistered medical images using cross-view transformers. In: International Conference on Medical Image Computing and Computer-Assisted Intervention. pp. 104--113. Springer (2021)

\bibitem{wang2020cross}
Wang, X., et~al.: Cross-batch memory for embedding learning. In: Proceedings of the CVPR. pp. 6388--6397 (2020)

\bibitem{wu2021cvt}
Wu, H., et~al.: Cvt: Introducing convolutions to vision transformers. In: Proceedings of the ICCV. pp. 22--31 (2021)

\bibitem{mem_vit}
Wu, Y., Rabe, M.N., Hutchins, D., Szegedy, C.: Memorizing transformers. In: International Conference on Learning Representations (2022), \url{https://openreview.net/forum?id=TrjbxzRcnf-}

\bibitem{xu2021co}
Xu, W., et~al.: Co-scale conv-attentional image transformers. In: Proceedings of the ICCV. pp. 9981--9990 (2021)

\bibitem{xue2022protopformer}
Xue, M., et~al.: Protopformer: Concentrating on prototypical parts in vision transformers for interpretable image recognition. arXiv preprint arXiv:2208.10431  (2022)

\bibitem{yuan2021incorporating}
Yuan, K., et~al.: Incorporating convolution designs into visual transformers. In: Proceedings of the ICCV. pp. 579--588 (2021)

\end{thebibliography}

\end{document}